%% file: world_events.tex
\documentclass[sigconf,natbib=true,anonymous=false]{acmart}

\AtBeginDocument{%
  \providecommand\BibTeX{{%
    \normalfont B\kern-0.5em{\scshape i\kern-0.25em b}\kern-0.8em\TeX}}}

\copyrightyear{2022}
\acmYear{2022}
\setcopyright{acmcopyright}
\acmConference[WSDM '22]{Proceedings of the Fifteenth ACM International Conference on Web Search and Data Mining}{February 21--25, 2022}{Tempe, AZ, USA}
\acmBooktitle{Proceedings of the Fifteenth ACM International Conference on Web Search and Data Mining (WSDM '22), February 21--25, 2022, Tempe, AZ, USA}
\acmPrice{15.00}
\acmDOI{10.1145/3488560.3498452}
\acmISBN{978-1-4503-9132-0/22/02}

\acmSubmissionID{wsdmfp406}

\graphicspath{ {Figures/} }

\usepackage{soul}
\usepackage{url}
\usepackage{amsmath}
\usepackage{booktabs}
\usepackage{varwidth}
\usepackage{amsfonts}
\usepackage{xspace}
\usepackage{enumitem}
\usepackage{array}
\usepackage{multirow}
\usepackage{caption}
\usepackage{subcaption}
\usepackage{tablefootnote}
\usepackage[export]{adjustbox}
\usepackage{xcolor}
\usepackage{amsthm}
\usepackage{todonotes}
\usepackage{graphicx}

\begin{document}

\fancyhead{}

%%
%% The "title" command has an optional parameter,
%% allowing the author to define a "short title" to be used in page headers.

\title{Leveraging World Events to Predict E-Commerce Consumer Demand under Anomaly}

%%
%% The "author" command and its associated commands are used to define
%% the authors and their affiliations.
%% Of note is the shared affiliation of the first two authors, and the
%% "authornote" and "authornotemark" commands
%% used to denote shared contribution to the research.

\author{Dan Kalifa}
\affiliation{
  \institution{Technion}
 \country{}
}
\email{kalifadan@cs.technion.ac.il}

\author{Uriel Singer}
\affiliation{
  \institution{Technion}
 \country{}
}
\email{urielsinger@cs.technion.ac.il}

\author{Ido Guy$^*$}
\affiliation{
  \institution{Ben-Gurion University of the Negev}
 \country{}
}
\thanks{$^*$Research was conducted while working at eBay}
\email{idoguy@acm.org}

\author{Guy D. Rosin}
\affiliation{
  \institution{Technion}
 \country{}
}
\email{guyrosin@cs.technion.ac.il}

\author{Kira Radinsky}
\affiliation{
  \institution{Technion}
 \country{}
}
\email{kirar@cs.technion.ac.il}

%% By default, the full list of authors will be used in the page
%% headers. Often, this list is too long, and will overlap
%% other information printed in the page headers. This command allows
%% the author to define a more concise list
%% of authors' names for this purpose.

\input{keywords}

\input{macros}
%%
%% The abstract is a short summary of the work to be presented in the
%% article.
\renewcommand{\shortauthors}{Kalifa et al.}
\begin{abstract}
\input{0-abstract}
\end{abstract}

%% A "teaser" image appears between the author and affiliation
%% information and the body of the document, and typically spans the
%% page.
% \begin{teaserfigure}
%   \includegraphics[width=\textwidth]{sampleteaser}
%   \caption{Seattle Mariners at Spring Training, 2010.}
%   \Description{Enjoying the baseball game from the third-base
%   seats. Ichiro Suzuki preparing to bat.}
%   \label{fig:teaser}
% \end{teaserfigure}

%%
%% This command processes the author and affiliation and title
%% information and builds the first part of the formatted document.
\maketitle
\renewcommand*{\thefootnote}{\arabic{footnote}}
\input{1-intro}
\input{2-rw}
\input{3-framework}
\input{4-setup}
\input{5-results}
\input{6-conclusions}

%%
%% The next two lines define the bibliography style to be used, and
%% the bibliography file.
\bibliographystyle{ACM-Reference-Format}
\bibliography{world_events}

%%
%% If your work has an appendix, this is the place to put it.
% \appendix
% \input{8-appendix}

\end{document}

%% file: keywords.tex
%%
%% The code below is generated by the tool at http://dl.acm.org/ccs.cfm.
%% Please copy and paste the code instead of the example below.
%%
\begin{CCSXML}
<ccs2012>
<concept>
<concept_id>10010147.10010257</concept_id>
<concept_desc>Computing methodologies~Machine learning</concept_desc>
<concept_significance>500</concept_significance>
</concept>
</ccs2012>
\end{CCSXML}

\ccsdesc[500]{Computing methodologies~Machine learning}

%%
%% Keywords. The author(s) should pick words that accurately describe
%% the work being presented. Separate the keywords with commas.
\keywords{world events; anomalies; e-commerce; forecasting}

%% file: macros.tex
\newcommand{\specialcell}[2][c]{%
  \begin{tabular}[#1]{@{}c@{}}#2\end{tabular}}

\newcommand{\us}[1]{\todo[inline]{US: #1}}
\newcommand{\dk}[1]{\todo[color=green!50, inline]{DK: #1}}
\newcommand{\kr}[1]{\todo[color=red!30, inline]{KR: #1}}
\newcommand{\ig}[1]{\todo[color=blue!40, inline]{IG: #1}}
\newcommand{\td}[1]{\todo[color=black!40, inline]{TO-DO: #1}}

\newcommand\footnoteref[1]{\protected@xdef\@thefnmark{\ref{#1}}\@footnotemark}

\newcommand{\ra}[1]{\renewcommand{\arraystretch}{#1}}

\newcommand{\argmin}{\mathop{\mathrm{argmin}}} 

%% Tables
\newcolumntype{L}[1]{>{\raggedright\let\newline\\\arraybackslash\hspace{0pt}}m{#1}}
\newcolumntype{C}[1]{>{\centering\let\newline\\\arraybackslash\hspace{0pt}}m{#1}}
\newcolumntype{R}[1]{>{\raggedleft\let\newline\\\arraybackslash\hspace{0pt}}m{#1}}
\newcommand\independent{\protect\mathpalette{\protect\independenT}{\perp}}
\def\independenT#1#2{\mathrel{\rlap{$#1#2$}\mkern2mu{#1#2}}}
\def\multitable#1{\multicolumn{1}{p{0.5cm}}{\centering #1}}

\def\PP{{\mathbb P}}
\def\EE{{\mathbb E}}
\def\cov{\text{cov}}

\def\G{\mathbf{G}}
\def\E{\mathbf{E}}
\def\V{\mathbf{V}}
\def\X{\mathbf{X}}
\def\Y{\mathbf{Y}}
\def\A{\mathbf{A}}
\def\h{\mathbf{h}}

\theoremstyle{definition}

\newcommand{\change}[1]{\color{red}{#1 }\color{black}}

%% file: 0-abstract.tex
Consumer demand forecasting is of high importance for many e-commerce applications, including supply chain optimization, advertisement placement, and delivery speed optimization. 
However, reliable time series sales forecasting for e-commerce is difficult, especially during periods with many anomalies, as can often happen during pandemics, abnormal weather, or sports events. Although many time series algorithms have been applied to the task, prediction during anomalies still remains a challenge. 
In this work, we hypothesize that leveraging external knowledge found in world events can help overcome the challenge of prediction under anomalies. We mine a large repository of 40 years of world events and their textual representations. 
Further, we present a novel methodology based on transformers to construct an embedding of a day based on the relations of the day’s events. 
Those embeddings are then used to forecast future consumer behavior. 
We empirically evaluate the methods over a large e-commerce products sales dataset, extracted from eBay, one of the world’s largest online marketplaces. We show over numerous categories that our method outperforms state-of-the-art baselines during anomalies.

%% file: 1-intro.tex
\section{Introduction}
\label{sec:intro}
\begin{figure}
    \centering
    \subfloat[\centering \small{Original time series.} \label{fig:football_sales}]{{\includegraphics[width=3.8cm]{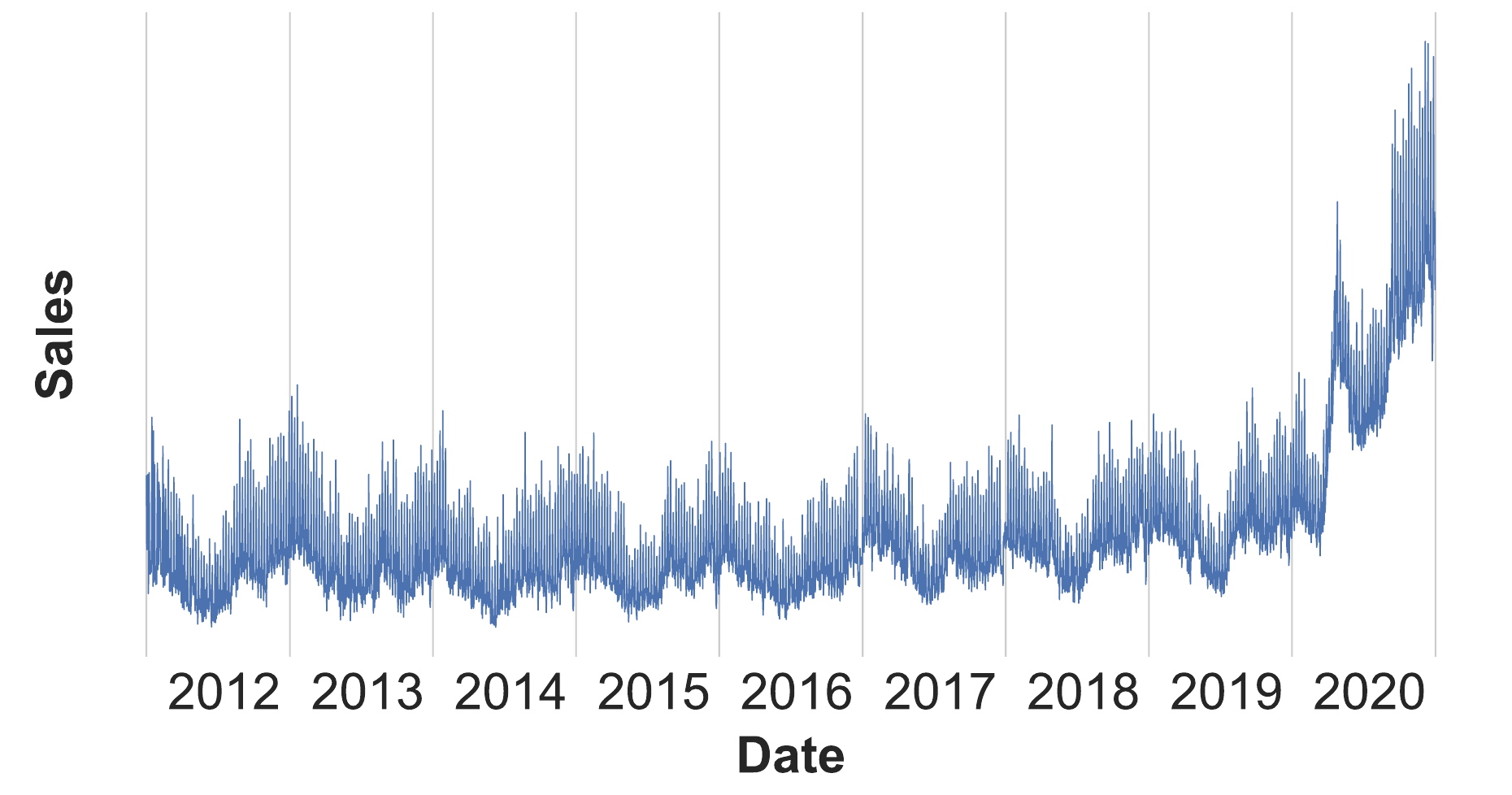} }}%
    \qquad
    \subfloat[\centering \small{The residual component.} \label{fig:football_residual}]{{\includegraphics[width=3.8cm]{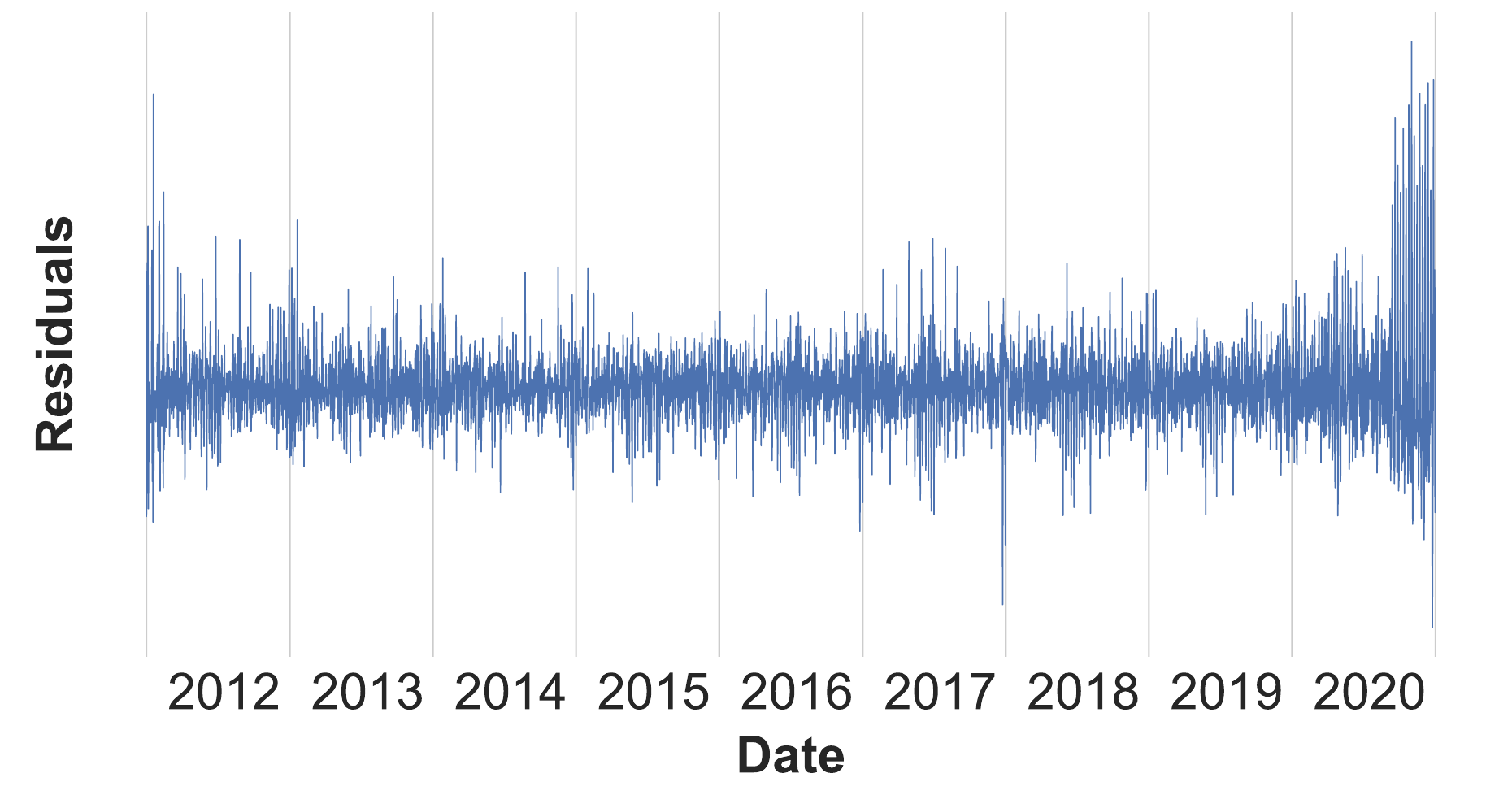} }}
    \vspace{-3mm}
    \caption{\small{Sales of the Football Cards category.}}
    \label{fig:sales_series}
\end{figure}

Demand forecasting is the process of predicting future customer demand, usually approximated through product sales. 
The quest for more accurate product sales forecasting is highly important in numerous e-commerce applications including supply chain optimization, advertisement placement, and delivery speed optimization. The latter is one of the dominating factors of user satisfaction during online purchasing~\cite{Sales_dfs_Qi2019}. Most approaches to the problem model it as a time series forecasting task~\cite{Sales_dfs_Qi2019, Ekambaram2020AttentionBM, Fattah2018ForecastingOD}. Classical time series models, such as ARIMA~\cite{Young1972TimeSA}, have been applied to the task and reached impressive results when the time series exhibited trend or seasonality~\cite{Omar2016AHN, Fattah2018ForecastingOD, Yermal2017ApplicationOA}. 
However, reliable time series forecasting for e-commerce is difficult, especially during high-variance periods (e.g., holidays, sporting events, pandemics), because event prediction is dependent on a variety of external factors, such as weather, health status, marketing trends, etc., all of which add to the forecast's uncertainty ~\cite{Zhu2017DeepAC, Sales_dfs_Qi2019}.
Due to its end-to-end modeling and ease of including exogenous variables ~\cite{Fischer2018DeepLW, Ruta2020DeepBL, Zhu2019ANA, Chimmula2020TimeSF}, time series modeling based on the Long Short Term Memory (LSTM) model has lately gained prominence ~\cite{Hochreiter1997LongSM}.
The LSTM network has been shown to be capable of simulating complex nonlinear feature interactions in time series~\cite{Liu2018StockTP, Sales_dfs_Qi2019}, which is important for simulating extreme events.
However, the problem of time series prediction during anomalies not explained by recurring exogenous variables remains an open challenge. 
Consider predicting the increase in sales of sports trading cards during COVID-19 (Figure~\ref{fig:football_sales}).
The COVID-19 pandemic has pushed many shoppers to make their purchases online as many physical retail outlets across the world are either closed due to lockdown measures or have limited capacity to maintain social distancing.
As shown in the figure, such increase in sales was not present in the past. The lack of history precludes the use of most state-of-the-art forecasting methods.

In this work, we present a methodology to leverage events as an auxiliary information for time series prediction during anomalies. 
We hypothesize that certain types of events increase or decrease future economic demand. 
We devise an event embedding model of a day, and leverage it to predict future economic behavior. 
To create a day's embedding, one might consider all the events occurring on that day. However, this representation might lead to spurious correlations when predicting future events, as the amount of events occurring every day is extremely large. 
Therefore, we attempt to identify a subset of the events to consider for that day's embedding.
Intuitively, we wish to filter out one-time events that don't usually occur with other events of that day. We therefore identify the set of highly-associated events occurring on that day.

To learn the associations between world events,
we mine 40 years of data (1980-2019). We focus on Wikipedia events, resulting in approximately 20,000 distinct events. 
As each specific event occurs only once, we wish to create an event-level embedding to allow generalization. 
We represent each event using Wikipedia2Vec \cite{Yamada2020Wikipedia2VecAE}, which embeds information about the content of the event, as well as Wikipedia's link graph.
But how should those event embeddings be aggregated to represent a day? How can we identify the highly-associated events representing that specific day?
To answer these questions, we present a novel adversarial encoder framework based on transformers. 
The encoder has two optimization tasks: (1) reconstruct the events of the day; and 
(2) learn the strength of association between the day's events. We apply masking and add an additional optimization task of reconstructing the masked events given the unmasked events. Intuitively, we attempt to evaluate, given the other day's events, which events would occur had it not been for the masked event.
The attention layers of the architecture allow to create the final day's embedding, while encapsulating the strength of the association between the events.
Once each day's embeddings are constructed, we apply a deep learner to learn the association between the day's embeddings and future product sales. 

We evaluate our work empirically over five different real-world product sales time series from eBay, in the years 2012--2020.  We focus on predictions during anomalies (i.e., the entire year of 2020) and show that leveraging the day's embedding learnt externally significantly boosts the results of the prediction.  We compare to state-of-the-art times series methods, including those that include exogenous events. We experiment with several architectures for prediction over the day's embeddings and show LSTMs reach the best results. 

The contributions of this work are threefold:
(1) We leverage a large repository of world events with their textual representations and present a novel methodology to construct an embedding of a day. 
The embedding is based on the strength of association of the day's events. The events' relations are learnt based on a novel transformer-based architecture, which learns to reconstruct the day while learning the association between its events. 
(2) We leverage the day embeddings to forecast future consumer behavior. To the best of our knowledge, this is the first work to successfully show the merit of leveraging world events to predict economic behavior during otherwise unpredicted anomalies.  
(3) We empirically evaluate the methods over a large e-commerce product dataset, extracted 
from eBay, one of the world’s largest online marketplaces.
We show over numerous categories that our method outperforms state-of-the-art baselines during anomalies. We contribute the code and data to the community for further research.

%% file: 2-rw.tex
\section{Related Work}
\label{sec:rw}

\paragraph{Time series forecasting}
Time series forecasting is an integral part of the automation and optimization of business processes, especially in retail and e-commerce~\cite{Sales_dfs_Qi2019, Faloutsos2019ForecastingBT}.
Several studies~\cite{Sales_dfs_Qi2019, Loureiro2018ExploringTU, Ekambaram2020AttentionBM, Zhao2017SalesFI, Bandara2019SalesDF, Singh2020SalesFF} focused on this task, using various forecasting methods, ranging from statistical ones to advanced deep learning models. 

One of the common forecasting approaches is the Autoregressive Integrated Moving Average (ARIMA) model~\cite{Young1972TimeSA}, which combines an autoregressive (AR) model and a moving average (MA) model. ARIMA has been applied in various domains, such as the food industry~\cite{Fattah2018ForecastingOD},
cryptocurrency industry~\cite{Yenidoan2018BitcoinFU},
and e-commerce~\cite{Sales_dfs_Qi2019}. 
AR models forecast future values using a linear combination of the past values, applying a regression of the variable against itself. 
Contrary to AR models, MA models predict future values based on the average of past observations, with an equal weight for all. Other MA models, like the exponential moving average, assign to past observations exponentially decreasing weights over time~\cite{Hyndman2013ForecastingPA}.

With the increasing popularity of deep learning methods, many neural network architectures have been applied to time series prediction and showed promising results~\cite{Sales_dfs_Qi2019, Brezak2012ACO}. 
One of the advantages of these models is the accurate results they achieved on both linear and nonlinear data, contrary to the statistical methods that have been tailored for linear data. 
Recurrent Neural Networks (RNNs) and in particular Long Short Term Memory Networks (LSTMs) were successfully applied for numerous time series forecasting tasks~\cite{Fischer2018DeepLW, Ruta2020DeepBL, Zhu2019ANA, Garg2021EvaluationOT, Chimmula2020TimeSF}.
The main reason for their superior performance is their ability to capture long-term dependencies~\cite{Hochreiter1997LongSM}, as often needed during time series prediction. 
Additional studies also attempted other architectures, such as convolutional neural networks (CNNs) ~\cite{Zhao2017SalesFI} and transformer-based
models~\cite{Zhou2021InformerBE}. Although these models can be used for time series forecasting, in this paper we focus on the relative improvement and therefore opted to use LSTM, which demonstrated state-of-the-are performance.

\paragraph{Sales forecasting}
Sales forecasting is a special case of time series forecasting. Many of the sales forecasting models address e-commerce scenarios~\cite{Sales_dfs_Qi2019, Zhao2017SalesFI, Bandara2019SalesDF, Singh2020SalesFF} and specific domains, such as Fashion~\cite{Loureiro2018ExploringTU, Ekambaram2020AttentionBM}.
Most of them apply deep learning models, which are able to predict future values based on external features alongside past values. Statistical models (e.g., ARIMA), on the other hand, have access to past values only. 
Recently, \citeauthor{Sales_dfs_Qi2019} proposed a Seq2Seq architecture (GRU) for product sales forecasting in e-commerce, exploiting heterogeneous sales-related features, proactive promotion campaigns, and competing relations between substitutable products. 
A more general framework to handle exogenous variables during time series prediction was presented by \citeauthor{Taylor2017ForecastingAS}. Their algorithm, Prophet~\cite{Taylor2017ForecastingAS}, is based on an additive model where non-linear trends are fit with seasonality and holidays or other recurrent events. It was applied in numerous studies and achieved state-of-the-art (SOTA) performance~\cite{Yenidoan2018BitcoinFU, Thiyagarajan2020ATF}.
Recently, NeuralProphet~\cite{Triebe2019ARNetAS, Taylor2017ForecastingAS}, a neural network-based extension of Prophet, reached SOTA results for several time series prediction tasks. 
In this work, we suggest a modeling that includes a wide variety of exogenous variables represented by world events, and present a model that learns to leverage their embedding for sales forecasting.

\paragraph{Prediction under anomalies}
Although time series forecasting models continue to improve, still, most models struggle to handle time series anomalies, i.e., points in time that exceed normal behavior (e.g., seasonality and trendiness) ~\cite{Geiger2020TadGANTS}. 
Most works until today handled the task of anomaly identification~\cite{Ghrib2020HybridAF, Maya2019dLSTMAN, Munir2019DeepAnTAD}, but the task of prediction during anomalies still remains a challenge~\cite{Munir2019DeepAnTAD}.
In this work, we attempt to improve forecasting during anomalies by leveraging world events.

\paragraph{Leveraging world events}
World events are a significant part of history.
The task of future event prediction has been tackled many times in recent years~\cite{radinsky2013mining,zhao2020event}, while leveraging world events was recently shown to be useful for various tasks~\cite{rosin2021event, zhao2017constructing, Taylor2017ForecastingAS}.
However, to the best of our knowledge, leveraging world events has not been applied to time series forecasting tasks, without hand-picking specific events. 
In this work, we leverage world events for product sales prediction. To overcome the fact that most events are one-time occurrences, thus prohibiting generalization, we use their textual descriptions and links from Wikipedia and create event embeddings. We then present a generative-adversarial model that attempts to understand deep relations between those events.

%% file: 3-framework.tex
\section{Proposed Method}
\label{sec:framework}
Let $c$ be a category of products, 
$S_t^c$ the product demand of category $c$ on day $t$ (i.e., the total number of items in this category sold that day), and
$E_t$ the set of all events that occurred on day $t$. 
Given a category $c$ and a day $t$, our goal is to predict future sales in a window of $W$ days: $S_{t}^c, ..., S_{t + W - 1}^c$, given historical product sales
 $S_{t-1}^c, S_{t-2}^c,..., S_{t-N}^c$ and world events $E_{t - 1}, ..., E_{t - N}$,
 where $N$ is the history length (in days).

In this section, we introduce \textit{GAN-Event}, a novel adversarial encoder framework based on transformers that is designed to learn the deep relations between world events.
We leverage adversarial training where our generator ($G$) and discriminator ($D$) are based on transformers~\cite{Vaswani2017AttentionIA} instead of the commonly used multilayer perceptrons ~\cite{Goodfellow2014GenerativeAN}. Using transformers allows us to receive an input of varying size of events, as the number of events per day can vary among days.

\begin{figure*}
\centering
    \subfloat[The generator model.]
        {
            \includegraphics[width=0.30\textwidth]{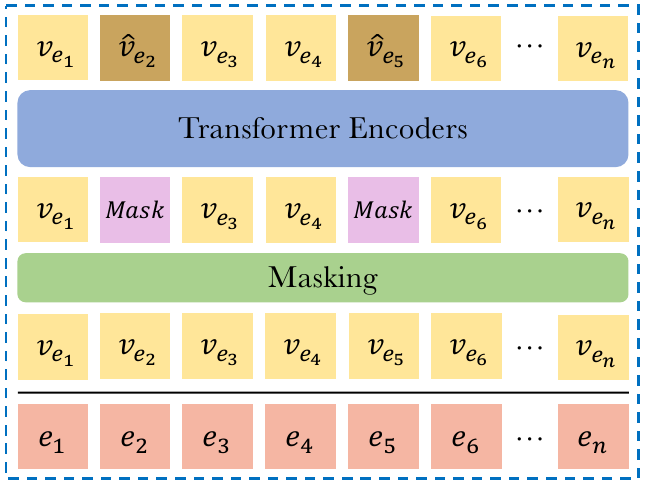}
            \label{fig:g_model}
        }
    \subfloat[The discriminator model.]
        {
            \includegraphics[width=0.285\textwidth]{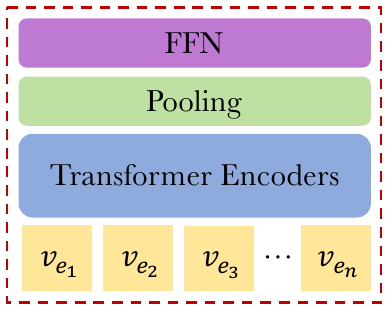}
            \label{fig:d_model}
        }
    \subfloat[The \textit{GAN-Event} model.]
        {  
            \includegraphics[width=0.26\textwidth]{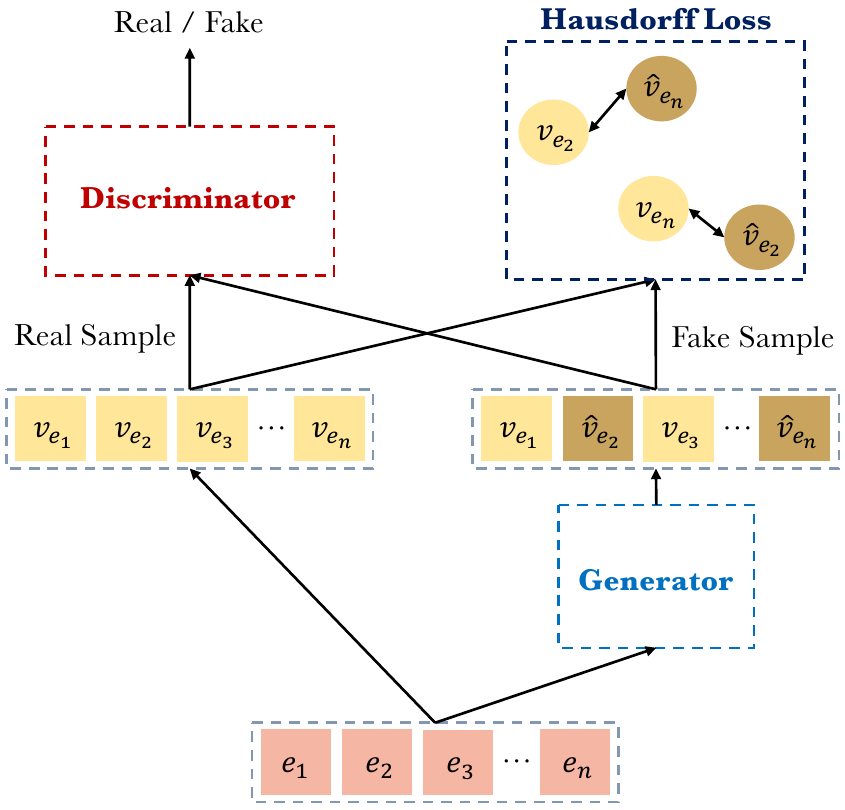}
            \label{fig:gan_model}
        }
    \vspace{-3mm}
    \caption{\small{The \textit{GAN-Event} architecture. (a) Given a set of a real day's events and their corresponding embedding vectors, the generator $G$ masks k\% of the events randomly, and attempts to reconstruct the masked events based on the unmasked events. (b) The discriminator $D$ learns the strength of association between the day's events. It receives a set of event embeddings and predicts whether the set of events is from a real day or not. (c) Given a set of a real day's events, \textit{GAN-Event} uses $G$ to reconstruct some of them (the events $G$ masked) and then executes $D$ to distinguish between the real day and $G$'s output. We use Hausdorff loss as an additional regulator that minimizes the distance between the reconstructed and original events.}}
\end{figure*}

Let $z \in \mathbb{R}^{n_t \cdot d}$ be a vector of event embeddings, where $n_t$ is the number of events on day $t$, and $d$ is the embedding size.
We define the generator $G(z; \theta_g)$ as a differentiable function 
$G: \mathbb{R}^{n_t \cdot d} \rightarrow \mathbb{R}^{n_t \cdot d}$ represented by a transformer encoder with parameters $\theta_g$. The function outputs a vector $z' \in \mathbb{R}^{n_t \cdot d}$ of generated event embeddings.
Additionally, we define the discriminator $D(z; \theta_d)$ to be a differentiable function $D: \mathbb{R}^{n_t \cdot d} \rightarrow \mathbb{R}$ represented by a transformer encoder with parameters $\theta_d$. The function outputs a single scalar, $D(z)$, representing the probability that $z$ came from the dataset (i.e., a real day's events) rather than the generator $G$.
We train $G$ to reconstruct the events of the day, i.e., generate events (Section~\ref{sec:Generator}), while the discriminator $D$ learns the strength of association between the day's events (Section~\ref{sec:Discriminator}). It aims to discriminate between a real day's events to generated 
ones. See Figure~\ref{fig:gan_model} for an illustration of the \textit{GAN-Event} architecture.
Finally, we present the \textit{GAN-Event LSTM} model (Section~\ref{sec:GAN_Event_LSTM}), which leverages the representation generated by the adversarial encoder for sales forecasting.

\subsection{GAN-Event Generator}
\label{sec:Generator}
The \textit{GAN-Event} generator (architecture illustrated in Figure~\ref{fig:g_model}) reconstructs events based on the context of other events. Its goal is to learn the distribution of world events in a day.
Intuitively, in order to learn this distribution, we need to answer the question: \textit{"Assuming events A, B, and C occurred on a certain day. Can we predict the occurrence of event B based on events A and C?"}.
If the answer is yes, we might conclude that event B probably has a strong relation to event A and event C occurrence. By answering this type of question, we can learn an approximation of the deep relations between events. Therefore, the generator has an optimization task to reconstruct events of a day, based on the context of other events occurring in that day.
Formally, for each day $t$ in our training set, the generator receives as input all the events which occurred that day. We also include the events of the day before it, since the effect of certain events is not immediate. Let these events be $E = E_t \cup E_{t-1} = \{e_1, e_2, ..., e_n\}$.
For each event $e \in E$ we create an embedding vector $v_e$ from Wikipedia2Vec~\cite{Yamada2020Wikipedia2VecAE}. Let these vectors be $V = \{v_{e_1}, v_{e_2}, ..., v_{e_n}\}$.
During training, the generator $G$ masks $k\%$ of all events.
By masking an event, we replace its embedding vector $v_e$ with a special mask event vector, which is a learned parameter of the model. We call the set of masked events $E_{mask} \subset E$, and the new vectors $V_{mask}$ (which is equal to $V$ with the masked vector replacing masked events). 
To reconstruct the masked events, we perform a forward pass through the generator's transformer encoders $\hat{V}=G(V_{mask})$. The outputs are the generated events, denoted by $\hat{V} = \{\hat{v}_{e_1}, \hat{v}_{e_2}, ..., \hat{v}_{e_n}\}$. 
To measure generation quality, i.e., measure the distance between the input vectors and generated vectors, we use cosine distance.
For each day $t$, we minimize the following reconstruction loss over the masked events:
\begin{equation}
\label{eq:reconstruct}
\min_{\theta_{g}} L_{rec} = \sum_{e \in E_{mask}} 1 - \cos(v_{e}, \hat{v}_{e}) 
\end{equation}
where $\cos$ is cosine similarity.

\subsubsection{Hausdorff Loss}
As the input of events has no special order, using the $L_{rec}$ loss defined above presents a problem: it is order-sensitive.
As shown in Figure~\ref{fig:g_model}, the generator attempts to reconstruct the masked event vectors. In case of generating the masked $e_2$ in the $5^{th}$ place, and the masked $e_5$ in the $2^{nd}$ place, the result is still fully correct, but $L_{gen}$ compares the events element-wise and so it would treat it as a mistake. We suggest using \textit{Hausdorff Distance} to solve this problem. 

\begin{definition}[Hausdorff Distance]
\label{def:sp}
Let $X$ and $Y$ be two non-empty subsets of a metric space $(M,d)$, where $d(a,B)=\inf \{ d(a,b) \mid b \in B \}$ is a distance function between a point $a$ and a subset $B$.
The Hausdorff distance between $X$ and $Y$ is defined by:
\begin{equation}
\label{eq:hausdorff}
d_{\mathrm H}(X,Y) = \max\left\{\,\sup_{x \in X} d(x,Y),\, \sup_{y \in Y} d(X,y) \,\right\}
\end{equation}
\end{definition}

Intuitively, this distance metric measures how far two subsets of a metric space are from each other without assuming any ordering. Two sets are close in the Hausdorff distance if there exists a matching between the two that minimizes the pairs distances. 

To overcome the drawback discussed above, we use the Hausdorff distance as the generator reconstruction loss $L_{rec}$. It measures the distance between the two spaces: the original event embeddings and the generated ones, without specifying event placement.
As the above metric is not differentiable, we replace the infimum function with the minimum function. Also, for smoothness, we replace the supremum and the maximum with the average function. Overall, we adapt the reconstruction loss presented in Eq.~\ref{eq:reconstruct} to:

\begin{equation}
\label{eq:hd_reconstruct}
\begin{split}
\min_{\theta_{g}} L_{rec} = \frac{1}{2} 
\Bigg[ 
\frac{1}{\lvert E_{mask} \rvert} \sum_{e \in E_{mask}} \min_{e' \in E_{mask}} d\left(v_{e}, \hat{v}_{e'}\right)
+ \\
\frac{1}{\lvert E_{mask} \rvert} \sum_{e \in E_{mask}}\min_{e' \in E_{mask}} d\left(\hat{v}_{e}, v_{e'}\right) 
\Bigg] 
\end{split}
\end{equation}
where $d(\cdot)$ can be any distance metric. We chose to use the cosine distance $d(x,y)=1 - \cos(x, y)$, and experiment with other functions in our experiments.

A loss based on the Hausdorff distance was presented and used over GANs, image segmentation and localization tasks~\cite{Li2021HausdorffGI, Karimi2020ReducingTH, whd-loss}. To the best of our knowledge, we are the first to use it as a generic framework that can use any distance function, such as $L_1$, $L_2$, or cosine distance. In addition, we are the first to use this kind of loss with unordered sequences in transformer-based models.

\subsection{GAN-Event Discriminator}
\label{sec:Discriminator}
The discriminator $D$'s task is to learn the strength of association between the day's events. Similarly to the generator, the discriminator model is also transformer-based.
See Figure~\ref{fig:d_model} for an illustration of its architecture. 
Given a set of events, the discriminator learns to differentiate between two scenarios: (1)
the given set of events belongs to a real day or (2) the given set of events contains generated events. 
Formally, given $V = \{v_{e_1}, v_{e_2}, ..., v_{e_n}\}$, our task is to predict 1 (real) or 0 (fake). 
We utilize the adversarial learning approach, and therefore, while the role of the discriminator is to discriminate between real and fake days, the generator tries to ``fool'' it, resulting in the following adversarial loss:
\begin{equation}
\begin{split}
\min_{\theta_{g}}\max_{\theta_{d}} L_{adv} = \EE_{v \sim V}\log\left(D\left(v\right)\right) \\ + \ \EE_{v \sim V_{mask}}\log\left(1-D\left(G\left(v\right)\right)\right)
\end{split}
\end{equation}

The final loss of the \textit{GAN-Event} is therefore:
$$\min_{\theta_{g}}\max_{\theta_{d}} \lambda_r \cdot L_{rec} + \lambda_d \cdot L_{adv}$$
where $\lambda_d, \lambda_r$ are parameters of the model, representing the weights of the reconstruction and adversarial losses, respectively.

\subsection{GAN-Event LSTM}
\label{sec:GAN_Event_LSTM}
Long Short-Term Memory (LSTM) networks showed significant performance on many real-world applications due to their ability to capture long-term dependencies~\cite{Karevan2020TransductiveLF}.
Therefore, we leverage \textit{GAN-Event} along with an LSTM model for sales forecasting.  
For every day $t$, the input of the model is a vector which combines the sales value and the day's embedding vector $v_t$. Our task is to predict the sales of the next $W$ days. 
To construct $v_t$, we wish to leverage not only the events of that day but also their relations.
We wish to approximate the strength of association between the events, in addition to the event representations. 
For that, we use the trained generator of \textit{GAN-Event} (Section~\ref{sec:Generator}): 
For each event $e \in E$, we mask its matching vector $v_e$ and unmask the other vectors. 
We pass all these vectors through the generator, which reconstructs the masked event vector as $\hat{v}_e$. 
If this event has a strong relation to the other events, the generator will succeed to reconstruct the right event. Therefore, if the association is low the reconstruction will be low as well.
We set $v_t$ to be the mean of $\hat{v}_e$ for every $e \in E$.

%% file: 4-setup.tex
\section{Experimental Setup}
\label{sec:experiments}

\subsection{Datasets}
\label{sec:datasets}
In this work, we used two real-world datasets. The first is a world-event dataset extracted from Wikipedia using DBpedia and Wikidata.\footnote{We publish our code and data: \url{https://github.com/kalifadan/GAN-Event}}
The second is an e-commerce product dataset, extracted from eBay, 
one of the world's largest online marketplaces.

\paragraph{World-event dataset}
\label{sec:event_dataset}
For the world-event dataset, we leveraged Wikipedia. We mined DBpedia and Wikidata to extract structured data from Wikipedia.
Specifically, we focused on entities from the top 14 categories of the DBpedia ontology class \textit{Event}.  
For each category, we mined entities with the corresponding event type from DBpedia (e.g., \textit{dbo:FootballMatch}), that have an associated Wikipedia entry and date of occurrence. 
Then, to validate our data and complete missing values, we mined the date of occurrence from the corresponding Wikidata entities. 
Finally, we filtered out all events with an invalid occurrence date, or out of our focus period of 1980 until 2020. 
After the pre-processing stage, our event dataset contained 16,766 world events.
To obtain embeddings for world events we can use several embedding sources. We chose to leverage Wikipedia2Vec~\cite{Yamada2020Wikipedia2VecAE}, which embeds information about the content of the event, as well as Wikipedia's link graph.
We used a pre-trained model of Wikipedia2Vec, which produces vector representations (dimension of 100) for words and entities in Wikipedia.

\paragraph{E-commerce dataset}
\label{sec:e_comm_dataset}
We used e-commerce data collected from the US site of eBay, and restricted the data to purchases of items in the US only. 
We focused on leaf categories within the site's category taxonomy as they provide the lowest granularity with sufficient purchase information. 
We examined the following categories that demonstrated high sales volumes during 2020: \textit{Football Cards}, \textit{Cell Phone Cases}, \textit{Disposable Face Masks}, \textit{Wrist Watches} and \textit{Men's Athletic Shoes}. These categories are spanning diverse domains in order to show novelty across multiple disciplines. These categories are ranging from extraordinarily anomalous categories to categories with fewer anomalies. 
In this paper, we are interested in evaluating the ability of our model to predict anomalies in a given signal. We consider each category as a time series of sales and attempt to define the points of anomalies.
A well-known approach~\cite{Taylor2017ForecastingAS} to identify anomalies in times series is to first calculate its residuals using ``Seasonal and Trend decomposition using Loess'' (STL)~\cite{Cleveland1990STLAS}, and then identify the high residuals to be anomalies.
The underlying idea of STL is to subtract the trend (moving average) and seasonality (average period signal) from the original time series, resulting in what is referred to as residuals (see Figure~\ref{fig:football_residual} for an illustration). 
Prior work considers these residuals as anomalies~\cite{Pincombea2007AnomalyDI, Ogata1989StatisticalMF}, and specifically we focus on high residual values to represent the time series anomalies.
Table~\ref{table:datasets_ebay} presents the average sales volume and residual for each category, where the values are divided into 4 buckets: Extremely High, High, Medium, and Low.
Consider, for example, the Disposable Face Masks category. All its daily sales values were nearly zero until December 2019, and then its sales peaked in a short period, due to the COVID-19 pandemic. Hence, this category is unusual and with a very limited short history.
Each category's time series is split into training and test periods: the training set is composed of 8 years of data in total (2012-2019), while the test set includes one year (2020).

\begin{table}[t]
\label{test_datasets}
\centering\small
\setlength{\tabcolsep}{0.30em}
\caption{\label{table:datasets_ebay} \small{E-commerce dataset characteristics: average of daily sale volume and residual. The values are divided into 4 categories: Extremely High, High, Medium, and Low.}}
\vspace{-3mm}
\begin{tabular}{@{}lccccc@{}}
\toprule

\multirow{2}{*}[-0.5\dimexpr \aboverulesep + \belowrulesep + \cmidrulewidth]{Category Name} 
& \multicolumn{2}{c}{Training Set (2012-9)}
& & \multicolumn{2}{c}{Test Set (2020)} \\ 
\cmidrule(lr){2-3}\cmidrule(r){5-6}

& Volume & Residual 
& & Volume & Residual \\
 
\midrule

Football Cards  
& Medium & Medium 
& & High  & High  \\

Cell Phone Cases   
& High & High 
& & High & Medium  \\

Disposable Face Masks   
& Low & Low 
& & High  & Extremely High  \\

Wrist Watches   
& Medium & Low  
& & Low & Low   \\

Men's Athletic Shoes   
& Medium & Low 
& & Medium  &  Low   \\

\bottomrule
\end{tabular}
\end{table}

\subsection{Baselines}
\label{sec:baselines}
We empirically compare our model to common and state-of-the-art time series forecasting models, ranging from statistical models to deep learning models:

\paragraph{ARIMA}
Auto Regressive Integrated Moving Average (ARIMA) is a model that combines an Autoregressive representation and a Moving Average, for time series forecasting. The ARIMA model combines the power of both and has demonstrated high performance in several time series analysis tasks~\cite{Omar2016AHN, Fattah2018ForecastingOD, Yermal2017ApplicationOA}. In our implementation, we use the \textit{Auto-ARIMA} model, which seeks to identify the optimal parameters for an ARIMA model, using the Akaike Information Criterion (AIC)~\cite{Bozdogan1987ModelSA}.

\paragraph{LSTM}
\label{sec:lstm_normal}
A recurrent neural network (RNN) with long short term memory (LSTM)~\cite{Hochreiter1997LongSM} that has only historical sales information, without any knowledge of world events. We follow the suggested parameters from the \textit{Darts} library~\footnote{\url{https://unit8co.github.io/darts/}}, setting the input chunk length to 365 and the dropout to 0.3.

\paragraph{Prophet}
\textit{Prophet} is a model for forecasting time series data based on an additive approach, where non-linear trends are fit with yearly, weekly, and daily seasonality, plus holiday effects~\cite{Taylor2017ForecastingAS}. 
In our implementation, we use Prophet's default parameters.
In addition, since our e-commerce dataset focuses on the US market, we include the US holidays as events integrated into the model.

\paragraph{Neural Prophet}
\textit{Neural Prophet} is a neural network method for time series forecasting, inspired by Prophet. It leverages Gradient Descent for optimization and models autocorrelation using AR-Net, which has been shown to bring performance improvements in prior work~\cite{Triebe2019ARNetAS}. We use Neural Prophet's default parameters, and include the US holidays as well.

\paragraph{Event Neural Prophet}
One of the advantages of the Prophet algorithm, as compared to pure statistical models, is its ability to include recurring events in their modeling. 
To evaluate the merit of the algorithmic framework presented in this work, as opposed to the merit of using the external set of events, we present a Neural Prophet variant that considers the same set of events as our model.
We add to Neural Prophet all world events with their occurrence time, and refer to the resulting model as \textit{Event Neural Prophet}.

\subsection{Evaluation Metrics}
\label{sec:metrics}
For performance evaluation, we use two common time series metrics -- 
Mean Absolute Error (MAE) and weighted Mean Absolute Percentage Error (wMAPE). These evaluation metrics have been used in many applications, including sales forecasting in e-commerce~\cite{Sales_dfs_Qi2019} and fashion demand forecasting~\cite{Ekambaram2020AttentionBM}.

The Mean Absolute Error (MAE) metric measures the mean absolute difference between the prediction values and the actual values. Therefore, lower MAE values indicate more accurate predictions. Given a predicted sales value $\hat{y_{i}}$ and a real sales value $y_{i}$ of each day $i$ in the test set $D$,
the MAE is defined by:
$$MAE = \frac{1}{\left| D \right|}\sum_{i \in D}\left | y_{i} - \hat{y_{i}} \right |$$ 

\citeauthor{Sales_dfs_Qi2019} suggested a variant of MAE, the weighted Mean Absolute Percentage Error (wMAPE) metric, which is better suited for e-commerce forecasting.
The wMAPE metric takes the magnitude of product sales into account by weighting the errors by the actual sales values:
$$wMAPE = \sum_{i \in D}\left | y_{i} - \hat{y_{i}} \right | / \sum_{i \in D}\left | y_{i} \right |$$
This metric is preferred for e-commerce forecasting since popular products with a high volume of sales should contribute more to the metric than unpopular products~\cite{Makridakis1993AccuracyMT, Sales_dfs_Qi2019}. 
We report MAE@K and wMAPE@K for $K{\in}{\{5,10,20\}}$, which evaluate models' performance in the K most anomalous days (i.e., with the largest residual values).

\subsection{Experimental Methodology}
\label{sec:exp_method}
In our experiments, given a category $c$ and day $t$ in the test set, our task is to predict future sales in a window of $W$ days, given a history of size $N$ days (Section~\ref{sec:framework}). We set $W=30$ (i.e., predict a full month ahead) and use all available past data (maximal $N$). Although our prediction granularity is one month, our test set contains one year (2020, see Section~\ref{sec:datasets}). 
Thus, we test on each of the 12 months of the test set, separately, after training for all preceding data (i.e., 2012-2019 plus the 2020 months leading up to the month aimed for prediction). After creating predictions for all months in 2020, we measure the MAE@K and wMAPE@K metrics over the K most anomalous days in 2020 (Section~\ref{sec:metrics}). 
We applied this evaluation approach since one month does not contain enough anomalous days to produce a meaningful evaluation, and for prediction, training should include the most recent preceding months.
We split the timeline in our datasets into train/validation/test sets. The test set is composed of the year 2020, and the remaining years are split 80\%:20\% between the training and validation sets, respectively. Then, we conduct a grid search to tune hyperparameters for all models, as reported below (Sections~\ref{sec:baselines}, \ref{sec:gan_event_impl}, \ref{sec:impact_model_arc}).

\subsection{GAN-Event LSTM Implementation Details}
\label{sec:gan_event_impl}
The generator $G$ and the discriminator $D$ have two transformer encoder layers, where each has four attention heads.
The generator's mask percentage $k$ is $25\%$, and the generator loss parameters are $\lambda_r=10$ and $\lambda_d=1$. 
After these layers there is an average pooling layer and three fully-connected layers with a Leaky ReLU between them, and finally a Sigmoid activation. 
Both models are trained for 100 epochs using a learning rate of 1e-4, weight decay of 1e-3, batches of size 32, and the AdamW optimizer~\cite{Loshchilov2019DecoupledWD}. Since the generator and the discriminator have different objective functions, each has its own optimizer.
Once the day embeddings are learned, we train an LSTM model for forecasting. 
For every day $t$, our input vector for the LSTM is a combination of the sales value (dimension of 1) and the day embedding vector $v_t$ (dimension of 100), as described in Section~\ref{sec:GAN_Event_LSTM}. 
We used the parameters from the \textit{Darts} library, setting the input chunk length to 365, hidden size to 404, and dropout of 0.3, to avoid overfitting. Further implementation details can be seen in our code.

%% file: 5-results.tex
\section{Results}
\label{sec:results}
Experimental results are reported on the five product categories summarized in Table~\ref{table:datasets_ebay}. 
We validate the statistical significance of the results with a permutation test~\cite{oden1975arguments} of a one-tailed paired Student's t-test for 95\% confidence. 
In all tables throughout the section, we boldface the best result for each category and metric, and use `$*$' to mark a statistically significant difference from the best result.

\subsection{Main Results}
\label{sec:main_res}
Table~\ref{table:main_results} presents the main results of our experiments, comparing our algorithm \textit{GAN-Event LSTM} with the five baseline methods.
We observe that the \textit{GAN-Event LSTM} significantly outperforms all baselines across the majority of categories and metrics. 

Inspecting the performance across categories, the gap between \textit{GAN-Event LSTM} and the other models is the largest on Football Cards, where the differences from other models are significant for all metrics. 
For Cell Phone Cases, all differences across metrics and $K$ values except one are significant. Next in term of performance significance are the Disposable Face Masks, followed by Wrist Watches, where the gap is clear and in many cases also significant. In contrast, in the Men’s Athletic Shoes category, all models perform similarly, without significant differences. \textit{GAN-Event LSTM} is still the best performing model for both metrics and $K{\in}\{5,10\}$.

Considering category anomaly levels (Table~\ref{table:datasets_ebay}), the most significant results of \textit{GAN-Event LSTM} are achieved on the most anomalous categories (i.e., medium, high, or extremely high levels), which are Football Cards, Cell Phone Cases, and Disposable Face Masks. 
In contrast, the Men’s Athletic Shoes category has a low level of anomalies, and indeed, its results do not demonstrate significant performance gaps when using the \textit{GAN-Event LSTM} model. 
On Wrist Watches, \textit{GAN-Event LSTM} performs relatively well, but 
the gap from other models is only partly significant. Overall, the \textit{GAN-Event LSTM} shows special strength in anomalous categories.  

At the model level, we make several observations regarding the performance of the baselines. 
First, Prophet, Neural Prophet, and Event Neural Prophet perform similarly across all categories. 
For 4 out of 5 categories, Event Neural Prophet outperforms the others, indicating the event information it uses has a slight yet consistent contribution to performance. 
The difference may be small due to the limited event information it considers (only name and date). Additionally, the vast majority of events are one-time events, which are worthless for Event Neural Prophet due to its inability to learn the effect of non-recurring events; as it considers only event names and dates, it cannot generalize to one-time events. 
The LSTM and ARIMA models reach second and third places, respectively, and perform similarly. These two models use only sales history as features, compared to the Prophet family models, which additionally consider holidays and recurring events. 
Nonetheless, the large gap between these models and the \textit{GAN-Event LSTM} emphasizes the importance of leveraging external knowledge found in world events for forecasting tasks, as we do in our proposed \textit{GAN-Event LSTM} model.
Our baseline results point to the superiority of the LSTM family over other models. 
This is aligned with previous work on e-commerce forecasting, where models based on various types of recurrent neural networks outperformed baselines: RNNs~\cite{Ekambaram2020AttentionBM}, GRUs~\cite{Sales_dfs_Qi2019}, and LSTMs~\cite{Bandara2019SalesDF}. 
The latter study also found that LSTM outperformed both ARIMA and Prophet, while ARIMA obtained better results than Prophet, similarly to our own results.

\begin{table}[t]
    \caption{\small{Main results of the sales forecasting task.}}
    \vspace{-3mm}
    \centering
    \scalebox{0.58}{

    \begin{tabular}{@{}llccccccccr@{}}\toprule
     Category & Model &    MAE@5 &  wMAPE@5 & MAE@10 & wMAPE@10 &  MAE@20 & wMAPE@20 \\
        
    \hline
        
    \multirow{6}{*}{Cards}
    
    & ARIMA &           6962* &            0.869* &            6867* &             0.927* &            5752* &             0.937* \\
    & Prophet &           8104* &            1.012* &            7529* &             1.017* &            6264*  &             1.020*  \\
    & Neural Prophet &           7971* &            0.995* &            7233* &             0.977* &            6012* &             0.979* \\
    & Event Neural Prophet &           7802* &            0.974* &            7126* &             0.962* &            5944*  &             0.968* \\
     & LSTM &           6453* &            0.806* &            6560* &             0.886* &            5478* &             0.892* \\
    & GAN-Event LSTM  &    \textbf{3239} &          \textbf{0.404} &        \textbf{4592} &    \textbf{0.620} &          \textbf{4012} &          \textbf{0.653} \\
        
    \hline
    
    \multirow{6}{*}{Cases}
    & ARIMA &           7410* &            0.845* &            6175* &             0.815* &            5329*  &             0.850*\\
    & Prophet &           8863* &            1.011* &            7699* &             1.016* &            6421* &             1.024*  \\
    & Neural Prophet &           9068* &            1.034* &            7731* &             1.020* &            6125*  &             0.977* \\
    & Event Neural Prophet &           8877* &            1.013* &            7027* &             0.927* &            5650* &             0.901* \\
    & LSTM &           7263 &            0.829 &            6120* &             0.808* &            5219* &             0.832* \\
    & GAN-Event LSTM  &      \textbf{5241} &        \textbf{0.598} &        \textbf{3960} &      \textbf{0.522} &        \textbf{3562} &         \textbf{0.568}  \\ 
    
    \hline
    
     \multirow{6}{*}{Masks}
    & ARIMA &          26275 &            1.002 &           19551* &             1.006* &           14010 &             1.005 \\
    & Prophet &          26929* &            1.027* &           20261* &             1.043* &           14409 &             1.034 \\
    & Neural Prophet &          27246* &            1.039* &           19424 &             0.999 &           14514 &             1.041 \\
    & Event Neural Prophet &          26113 &            0.996 &           19776* &             1.018* &           14136 &             1.014 \\
    & LSTM &          26284 &            1.002 &           19502* &             1.003* &           13972 &             1.002 \\
    & GAN-Event LSTM  &  \textbf{22166} &           \textbf{0.845} &          \textbf{16766} &      \textbf{0.863} &         \textbf{12732} &           \textbf{0.913}  \\
    
    \hline
    
    \multirow{6}{*}{Watches}
    &  ARIMA &           1386* &            0.868* &            1245* &             0.864* &            1121* &             0.892* \\
    &  Prophet &           1591* &            0.997* &            1491* &             1.035* &            1318*  &             1.049*  \\
    &  Neural Prophet &           1510 &            0.946 &            1397* &             0.970* &            1260* &             1.003* \\
    &  Event Neural Prophet &           1491 &            0.934 &            1357* &             0.941* &            1210* &             0.963* \\
    & LSTM  &           1394* &            0.873* &            1268* &             0.880* &            1102* &             0.878* \\
    &  GAN-Event LSTM  &       \textbf{1006} &           \textbf{0.630}    &        \textbf{844}   &     \textbf{0.586} &        \textbf{816} &            \textbf{0.649}  \\

    \hline
   
    \multirow{6}{*}{Shoes}
    &  ARIMA  &          13063 &            1.000 &            9483 &             1.007 &            6351 &             0.990 \\
    &  Prophet &          13079 &            1.001 &            9427 &             1.001 &            6427 &             1.001 \\
    & Neural Prophet &          13043 &            0.998 &            9440 &             1.003 &            6238 &             \textbf{0.972} \\
    & Event Neural Prophet &          13021 &            0.997 &            9523 &             1.011 &            \textbf{6236} &             \textbf{0.972} \\
    & LSTM &          13044 &            0.999 &            9451 &             1.004 &            6434 &             1.003 \\
    &  GAN-Event LSTM  &     \textbf{12851} &          \textbf{0.984} &         \textbf{9231} &      \textbf{0.980} &         6272 &           0.977  \\

    \bottomrule
    \end{tabular}
    }
    \label{table:main_results}
\end{table}

% ------------------------------------------------------------------------------------

% ------------------------------------------------------------------------------------
% ------------------------------- Ablation Tests -------------------------------------
% ------------------------------------------------------------------------------------
\subsection{Impact of Events Embedding}
In this section, we study different event embedding methods and their effect.
We leverage the LSTM model as the forecasting method as it reached the best performance among our baselines (Table~\ref{table:main_results}).

\paragraph{Event LSTM}
We suggest a baseline where we consider all events without any knowledge about their associations.  Instead of having a GAN generating event representations thus dissociating some of the non-association relations, we use average pooling to aggregate all input event vectors into a single vector of size $d$. 
Once the embedding is created, similar to what was performed in GAN-Event LSTM, we concatenate the sales value, creating a vector with dimension $d + 1$, which is then used as input to the LSTM network.

\paragraph{Weighted Event LSTM}
One might hypothesize that only large events have impact on forecasting. We therefore suggest a baseline which weighs each event based on its impact on the world. Similar to \textit{Event LSTM}, we disregard the underlying association between the events, and apply a weighted average as a pooling method, where the weight of each input event vector is based on the number of links of the event's Wikipedia entry. 
\vspace{2mm}

Table~\ref{table:ablation1} shows the comparison results. 
We can see that the \textit{GAN-Event LSTM} outperforms all LSTM baselines, consistently across all categories and metrics except for two metrics in Shoes which are not statistically significant. Across other categories, most differences are statsitically significant. The performance gap is especially large in anomalous categories (i.e., Cards, Cases, and Watches). Yet, the event-based baselines are outperformed by the plain LSTM, which predicts based on previous sales information only, across the majority of categories and metrics. 
This indicates that the way external event information is being incorporated into the model is of high importance. 
In the Event LSTM model, simple pooling methods do not appear to capture complex relations between events and generate noise instead of meaningful features. It can also be observed that pooling by plain average is preferable to weighted average across most categories and metrics. 
Overall, these observations reinforce the value of our \textit{GAN-Event}, which effectively captures meaningful event information.

\begin{table}
    \caption{\small{Comparison of LSTM models.}}
    \vspace{-3mm}
    \centering
    \scalebox{0.92}{
    \footnotesize
    \setlength{\tabcolsep}{0.23em}
    \begin{tabular}{@{}llccccccccr@{}}\toprule
    \multirow{2}{*}[-0.5\dimexpr \aboverulesep + \belowrulesep + \cmidrulewidth]{Category} &  \multirow{2}{*}[-0.5\dimexpr \aboverulesep + \belowrulesep + \cmidrulewidth]{Model} & \multicolumn{2}{c}{K=5} &  \multicolumn{2}{c}{K=10} & \multicolumn{2}{c}{K=20}\\
    \cmidrule(lr){3-4}   \cmidrule(lr){5-6}   \cmidrule(lr){7-8}
    & & MAE & wMAPE & MAE & wMAPE & MAE & wMAPE \\
    \midrule

    \multirow{4}{*}{Cards}
    
    & LSTM &           6453* &            0.806*  &            6560*  &             0.886*  &            5478*  &             0.892*  \\
    & Event LSTM &           7634*  &            0.953*  &            7196*  &             0.972*  &            5910*  &             0.962*  \\
    & Weighted Event LSTM &           7621*  &            0.951*  &            7468*  &             1.008*  &            6137*  &             0.999*  \\ 
    & GAN-Event LSTM  &      \textbf{3239} &          \textbf{0.404} &        \textbf{4592} &    \textbf{0.620} &          \textbf{4012} &          \textbf{0.653} \\
    
    \hline
    
    \multirow{4}{*}{Cases}
    & LSTM &             7263  &            0.829  &            6120*  &             0.808*  &            5219*  &             0.832*  \\
     & Event LSTM &           11380*  &            1.298*  &            9272*  &             1.224*  &            6758*  &             1.077*  \\
     & Weighted Event LSTM &          12714*  &            1.450*  &           10239*   &             1.351*  &            7607*  &             1.213*  \\
    & GAN-Event LSTM  &       \textbf{5241} &        \textbf{0.598} &        \textbf{3960} &      \textbf{0.522} &        \textbf{3562} &         \textbf{0.568}  \\ 

    \hline
    
    \multirow{4}{*}{Masks}
    & LSTM &            26284 &            1.002 &           19502* &             1.003* &           13972 &             1.002 \\
     & Event LSTM &          27134 &            1.035 &           19008 &             0.978 &           14603 &             1.047 \\
    & Weighted Event LSTM &          27836 &            1.062 &           18677 &             0.961 &           14099 &             1.011 \\
    & GAN-Event LSTM  &   \textbf{22166} &           \textbf{0.845} &          \textbf{16766} &      \textbf{0.863} &         \textbf{12732} &           \textbf{0.913}  \\
        
    \hline

    \multirow{4}{*}{Watches}
    & LSTM &             1394*  &            0.873*  &            1268*  &             0.880*  &            1102*  &             0.878*  \\
     & Event LSTM &            1549 &            0.970 &            1204*  &             0.836*  &             973 &             0.774 \\
     & Weighted Event LSTM &           1566*  &            0.981*  &            1298*  &             0.901*  &            1024*  &             0.815*  \\
    & GAN-Event LSTM  &        \textbf{1006} &           \textbf{0.630}    &        \textbf{844}   &     \textbf{0.586} &        \textbf{816} &            \textbf{0.649}  \\
    \hline

    \multirow{4}{*}{Shoes}
    & LSTM &            13044 &            0.999 &            9451 &             1.004 &            6434 &             1.003 \\
    & Event LSTM &          13035 &            0.998 &            9422 &             1.001 &            6512 &             1.015 \\
    & Weighted Event LSTM & \textbf{12441} &   \textbf{0.952} &            9257 &             0.983 &            6550 &             1.021 \\
    & GAN-Event LSTM  &     12851 &         0.984  &         \textbf{9231} &      \textbf{0.980} &         \textbf{6272} &           \textbf{0.977}  \\

    \bottomrule
    \end{tabular}
    }
    \label{table:ablation1}
    
\end{table}

% ------------------------------------------------------------------------------------
% ------------------------------- Ablation - CNN -------------------------------------
% ------------------------------------------------------------------------------------
\subsection{Impact of Model Architecture}
\label{sec:impact_model_arc}
The \textit{GAN-Event LSTM} is an architecture based on an LSTM. Another common architecture for time series forecasting is convolutional neural networks (CNN). In this section, we compare our \textit{GAN-Event LSTM} with a \emph{GAN-Event CNN}, an implementation based on a CNN architecture.
We use a variant of CNN, named temporal convolutional networks (TCN)~\cite{Bai2018AnEE}, which is an adaptation of the CNN architecture for time series prediction. A TCN consists of dilated, causal one-dimensional convolutional layers with the same input and output lengths. We reported the model's parameters in our code.
Table~\ref{table:ablation2} summarizes the comparison results over our test set. It can be seen that \textit{GAN-Event LSTM} outperforms \emph{GAN-Event CNN} across all categories and metrics, except for Men's Athletic Shoes. In this category, both models perform similarly and do not achieve good results, as their wMAPE is near to 1.
A similar finding, indicating that a CNN architecture performs substantially worse in forecasting tasks in e-commerce, was reported in~\cite{Sales_dfs_Qi2019}. LSTMs proved to be effective and scalable for several learning problems related to sequential data~\cite{Greff2017LSTMAS}, and it is therefore not surprising they are superior to CNNs in sales forecasting.

\begin{table}
    \caption{\small{Comparison of GAN-Event architectures.}}
    \vspace{-3mm}
    \centering
    \scalebox{0.92}{
    \footnotesize
    \setlength{\tabcolsep}{0.36em}
    \begin{tabular}{@{}llccccccccr@{}}\toprule
    \multirow{2}{*}[-0.5\dimexpr \aboverulesep + \belowrulesep + \cmidrulewidth]{Category} &  \multirow{2}{*}[-0.5\dimexpr \aboverulesep + \belowrulesep + \cmidrulewidth]{Model} & \multicolumn{2}{c}{K=5} &  \multicolumn{2}{c}{K=10} & \multicolumn{2}{c}{K=20}\\
    \cmidrule(lr){3-4}   \cmidrule(lr){5-6}   \cmidrule(lr){7-8}
    & & MAE & wMAPE & MAE & wMAPE & MAE & wMAPE \\
    \midrule

    \multirow{2}{*}{Cards}
    & GAN-Event CNN &  7982*  & 0.996*  & 7361*  & 0.994*  & 6127*  & 0.998*  \\
    & GAN-Event LSTM &     \textbf{3239} &          \textbf{0.404} &        \textbf{4592} &    \textbf{0.620} &          \textbf{4012} &          \textbf{0.653} \\
    \hline
    
    \multirow{2}{*}{Cases}
    & GAN-Event CNN & 8809 & 1.005 & 7269*  & 0.959*  & 5892  & 0.939   \\
    & GAN-Event LSTM &      \textbf{5241} &        \textbf{0.598} &        \textbf{3960} &      \textbf{0.522} &        \textbf{3562} &         \textbf{0.568}  \\ 
    \hline
    
    \multirow{2}{*}{Masks}
    & GAN-Event CNN & 26586 & 1.014 & 19678 & 1.013 & 13966 & 1.002  \\
    & GAN-Event LSTM &  \textbf{22166} &           \textbf{0.845} &          \textbf{16766} &      \textbf{0.863} &         \textbf{12732} &           \textbf{0.913}  \\
    \hline
    
    \multirow{2}{*}{Watches}
    & GAN-Event CNN & 1610*  & 1.009*  & 1467*  & 1.018*  & 1273*  & 1.013*  \\   
    & GAN-Event LSTM &      \textbf{1006} &           \textbf{0.630}    &        \textbf{844}   &     \textbf{0.586} &        \textbf{816} &            \textbf{0.649}  \\
    \hline

    \multirow{2}{*}{Shoes}
    & GAN-Event CNN &  12890 & 0.987 &  \textbf{9136} &  \textbf{0.970} &  \textbf{6230} &  \textbf{0.971}   \\
    & GAN-Event LSTM &     \textbf{12851} &          \textbf{0.984} &        9231 &      0.980 &         6272 &          0.977  \\

    \bottomrule
    \end{tabular}
    }
    \label{table:ablation2}
\end{table}

% ------------------------------------------------------------------------------------
% ------------------------------- Ablation - Losses ----------------------------------
% ------------------------------------------------------------------------------------
\subsection{Impact of GAN Loss Functions}
In this section, we compare several distance functions and evaluate their effect on the \textit{GAN-Event LSTM}'s performance.
While Hausdorff is used as a loss function, notice the function $d(\cdot)$ defined in Eq.~\ref{eq:hd_reconstruct} can be any distance function.
Therefore, we compare the cosine distance used in our model (Section~\ref{sec:Generator}) to $L_1$ and $L_2$ norms, two commonly used distance metrics~\cite{Isola2017ImagetoImageTW}. 
Using these metrics, we define the following $d(\cdot)$ functions, where x and y are vectors:
$d_{cosine}(x,y) = 1 - cos(x, y)$, $d_{L_1}(x,y) = \left| x - y \right|$, $d_{L_2}(x,y) = \left(x - y \right)^2$.
Inspired by natural language processing methods~\cite{kenton2019bert}, an additional loss function that could have been considered is the Categorical Cross-Entropy loss. This loss predicts the correct word out of a given vocabulary, and thus it is inappropriate for our setup; note that world events do not form a closed vocabulary (considering yet-to-be-known future events) and their representation is continuous (in the $\mathbb{R}^{d}$ space). 
Table~\ref{table:ablation4} summarizes the comparison results.
We can observe that cosine distance performs best as the distance function in the Hausdorff loss, for the majority of metrics across all categories. 
This observation can be explained by the nature of the world event embeddings; these are created by Wikipedia2Vec~\cite{Yamada2020Wikipedia2VecAE}, which also applied the cosine metric between entities (e.g., events) in its optimization process.

\begin{table}
    \caption{\small{Comparison Hausdorff loss distance functions.}}
    \vspace{-3mm}
    \centering
     \scalebox{0.92}{
    \footnotesize
    \setlength{\tabcolsep}{0.44em}
    \begin{tabular}{@{}llccccccccr@{}}\toprule
    \multirow{2}{*}[-0.5\dimexpr \aboverulesep + \belowrulesep + \cmidrulewidth]{Category} &  \multirow{2}{*}[-0.5\dimexpr \aboverulesep + \belowrulesep + \cmidrulewidth]{Function} & \multicolumn{2}{c}{K=5} &  \multicolumn{2}{c}{K=10} & \multicolumn{2}{c}{K=20}\\
    \cmidrule(lr){3-4}   \cmidrule(lr){5-6}   \cmidrule(lr){7-8}
    & & MAE & wMAPE & MAE & wMAPE & MAE & wMAPE \\
    \midrule
        
    \multirow{3}{*}{Cards}
    & L1  &      3878 &          0.484 &        5151 &     0.696 &        4346 &        0.708  \\
    & L2 &   3359 & 0.419 & 4797 & 0.648 & 4043 & 0.658 \\
    & Cosine  &      \textbf{3239} &          \textbf{0.404} &        \textbf{4592} &    \textbf{0.620} &          \textbf{4012} &          \textbf{0.653} \\
    \hline
    
    \multirow{3}{*}{Cases}
    & L1  &      5746  &        0.656  &        5447  &     0.719  &         4952*    &        0.790*   \\
    & L2 &       6023  & 0.687  & 5967*  & 0.787*  & 4690*  & 0.748*  \\
    & Cosine  &    \textbf{5241} &        \textbf{0.598} &        \textbf{3960} &      \textbf{0.522} &        \textbf{3562} &         \textbf{0.568}  \\  
    \hline
    
     \multirow{3}{*}{Masks}
    & L1  &     \textbf{22045} &         \textbf{0.841} &          16986 &     0.874 &         12895 &       0.925  \\
    & L2 &      24392 & 0.930 & 17841 &  0.918  & \textbf{12308} & \textbf{0.883}  \\
    & Cosine  &        22166 &           0.845 &          \textbf{16766} &  \textbf{0.863} &   12732 &     0.913  \\
    \hline
    
    \multirow{3}{*}{Watches}
    & L1  &      1039 &           0.651 &          919 &     0.638 &          \textbf{766} &          \textbf{0.610} \\
    & L2 &     1595* & 1.000* & 1107 & 0.768 & 875 & 0.697 \\
    & Cosine  &        \textbf{1006} &           \textbf{0.630}    &        \textbf{844}   &     \textbf{0.586} &         816 &            0.649  \\
    \hline
    
    \multirow{3}{*}{Shoes}
    & L1  &       12762 &         0.977 &        \textbf{9038} &    \textbf{0.960} &        \textbf{6211} &         \textbf{0.968}  \\
    & L2 &       \textbf{12656} & \textbf{0.969} & 9127 & 0.969 & 6240 & 0.972 \\
    & Cosine  &       12851 &          0.984 &         9231 &      0.980 &         6272 &           0.977  \\
    \bottomrule
    \end{tabular}
    }
    \label{table:ablation4}
 \end{table}

% ------------------------------------------------------------------------------------
% ------------------------------- Ablation - GAN -------------------------------------
% ------------------------------------------------------------------------------------
\subsection{Ablation Tests}
In this section, we set out to further explore the contribution of the \textit{GAN-Event LSTM}'s components, and report the results in Table~\ref{table:ablation3}.
We experiment with a version of GAN-Event LSTM without the discriminator ($\lambda_d = 0$), and then with a model without the reconstruction loss ($\lambda_r=0$). It can be seen that both components of the architecture have a significant impact on performance. 
As an additional ablation test, we consider removing the Hausdorff loss, i.e., leveraging the reconstruction loss presented in Eq.~\ref{eq:reconstruct}, rather than the Hausdorff Loss presented in Eq.~\ref{eq:hd_reconstruct}. We observe that removing the Hausdorff wrapper degrades performance significantly in 2 out of 5 categories, and insignificantly in other categories.

\begin{table}
    \caption{\small{Performance when excluding key model components.}}
    \vspace{-3mm}
    \centering
    \scalebox{0.92}{
    \footnotesize
    \setlength{\tabcolsep}{0.26em}
    \begin{tabular}{@{}llccccccccr@{}}\toprule
    \multirow{2}{*}[-0.5\dimexpr \aboverulesep + \belowrulesep + \cmidrulewidth]{Category} &  \multirow{2}{*}[-0.5\dimexpr \aboverulesep + \belowrulesep + \cmidrulewidth]{Model} & \multicolumn{2}{c}{K=5} &  \multicolumn{2}{c}{K=10} & \multicolumn{2}{c}{K=20}\\
    \cmidrule(lr){3-4}   \cmidrule(lr){5-6}   \cmidrule(lr){7-8}
    & & MAE & wMAPE & MAE & wMAPE & MAE & wMAPE \\
    \midrule
        
    \multirow{4}{*}{Cards}
    & GAN-Event LSTM &  \textbf{3239} &          \textbf{0.404} &        \textbf{4592} &    \textbf{0.620} &          \textbf{4012} &          \textbf{0.653} \\
    & $-$ Discriminator &  7636*  & 0.953*  & 7412*  & 1.001* & 6042*  & 0.984*   \\
    & $-$ Reconstruction loss &  6623*  & 0.827*  & 6573*  & 0.888*  & 5346*  & 0.871*   \\
    & $-$ Hausdorff wrapper &   5256*  & 0.656*  & 5884*  & 0.795*  & 5071*  & 0.826*   \\
    \hline

    \multirow{4}{*}{Cases}
    & GAN-Event LSTM &      \textbf{5241} &        \textbf{0.598} &        \textbf{3960} &      \textbf{0.522} &        \textbf{3562} &         \textbf{0.568}  \\ 
    & $-$ Discriminator* &   11856*  & 1.353*  & 9192*  & 1.213*  & 6478*  & 1.033*   \\
    & $-$ Reconstruction loss &  7882*  & 0.899*  & 7149*  & 0.943*  & 5741*  & 0.915*   \\
    & $-$ Hausdorff wrapper &    6897*  & 0.787*  & 5844*  & 0.771*  & 4902*  & 0.782*    \\
    \hline

    \multirow{4}{*}{Masks}
    & GAN-Event LSTM &  \textbf{22166} &           \textbf{0.845} &          \textbf{16766} &      \textbf{0.863} &         12732 &           0.913  \\
    & $-$ Discriminator &    27414* & 1.046* & 16921 & 0.871 & 13175 & 0.945  \\
    & $-$ Reconstruction loss &   23968 & 0.914 & 16903 & 0.870 & \textbf{12398} & \textbf{0.889}  \\
    & $-$ Hausdorff wrapper &     22851 & 0.871 & 16972 & 0.873 & 12552 & 0.900   \\
    \hline

    \multirow{4}{*}{Watches}
    & GAN-Event LSTM &     1006 &         0.630    &      844   &   0.586  &  816 &           0.649  \\
    & $-$ Discriminator &   1068 & 0.669 & 957* & 0.664* & 806 & 0.641  \\
    & $-$ Reconstruction loss &    864 & 0.541 & \textbf{659} & \textbf{0.458} & \textbf{644} & \textbf{0.513}  \\
    & $-$ Hausdorff wrapper &   \textbf{842} & \textbf{0.527} & 790 & 0.549 & 726 & 0.578  \\
    \hline
    
    \multirow{4}{*}{Shoes}
    & GAN-Event LSTM &    12851 &          0.984  &    9231 &  0.980  &  \textbf{6272} &           \textbf{0.977}  \\
    & $-$ Discriminator &   13133 & 1.005 & 9692 & 1.029 & 6310 & 0.983  \\
    & $-$ Reconstruction loss &     12998 & 0.995 & 9242 & 0.982 & 6322 & 0.985  \\
    & $-$ Hausdorff wrapper &     \textbf{12743} & \textbf{0.975} & \textbf{9121} & \textbf{0.969} & 6303 & 0.982   \\
    
    \bottomrule
    \end{tabular}
    }
    \label{table:ablation3}

\end{table}

%% file: 6-conclusions.tex
\section{Conclusions}
\label{sec:conclusions}
In this paper, we presented a novel method of leveraging external knowledge found in world events for sales forecasting during anomalies in e-commerce. 
We introduced a transformer-based architecture that learns to create an embedding of a day based on its events. 
Our approach leverages world events and their textual representations, extracted from Wikipedia.
It extracts an approximation of the association between the day’s events. This approximation, together with the events' text embeddings, are leveraged in a transformer-based encoder to create a day's embedding. 
These days' embeddings are integrated into an LSTM model to forecast future consumer behavior, a critically important task in numerous e-commerce applications and systems.
We empirically evaluate our method over a large e-commerce product sales dataset from eBay. 
We compare our model to the SOTA approaches for time series prediction, including models that consider events, but not their textual representations, and show significant statistical gains. We hypothesize that these performance gains stem from deeper semantic understanding of the events themselves. This understanding allows generalization for never-before seen events and estimating their impact on future sales. 
We show that learning the events' association as compared to merely leveraging all the day's events semantics has a significant impact on performance.
We perform several ablation tests and study the need for each part of the architecture. 
To the best of our knowledge, this work is the first to successfully show the merit of leveraging world events and their semantics to predict economic behavior during otherwise unpredicted anomalies. 

\newpage
\clearpage
\newpage